\documentclass[11pt,a4paper]{article}

\usepackage[utf8]{inputenc}
\usepackage[T1]{fontenc}
\usepackage[margin=1in]{geometry}
\usepackage{hyperref}
\usepackage{url}
\usepackage{booktabs}
\usepackage{multirow}
\usepackage{graphicx}
\usepackage{amsmath}
\usepackage{xspace}
\usepackage{natbib}
\usepackage{parskip}
\usepackage{titlesec}
\usepackage{caption}
\usepackage{subcaption}
\usepackage[table]{xcolor}
\usepackage{threeparttable}

\hypersetup{
  colorlinks=true,
  linkcolor=blue!60!black,
  citecolor=blue!60!black,
  urlcolor=blue!60!black
}

\newcommand{\bleu}{\textsc{BLEU}\xspace}
\newcommand{\chrf}{\textsc{chrF++}\xspace}
\newcommand{\ter}{\textsc{TER}\xspace}
\newcommand{\comet}{\textsc{COMET}\xspace}
\newcommand{\bertscore}{\textsc{BERTScore}\xspace}

\title{\textbf{Evaluating Large Language Models for Hausa and Fongbe\\
Machine Translation: Benchmarks, Failures, and Metric Reliability}}

\author{
  Mahounan Pericles Adjovi$^{1}$ \quad
  Roald Eiselen$^{2}$\quad
  Prasenjit Mitra$^{1}$\\[4pt]
  $^{1}$Carnegie Mellon University Africa, Kigali, Rwanda\\
  $^{2}$North-West University, South Africa\\[4pt]
  \texttt{madjovi@alumni.cmu.edu}
  \texttt{Roald.Eiselen@nwu.ac.za}
  \texttt{prasenjm@andrew.cmu.edu}
}

\date{}

\begin{document}

\maketitle

\begin{abstract}
We investigate the translation quality of current large language models (LLMs) for English-to-Hausa and English-to-Fongbe—two typologically distinct West African languages from the Afroasiatic and Niger-Congo families respectively—and evaluate whether standard automatic metrics reliably reflect human judgment for these low-resource languages. We evaluate four models (GPT-4o Mini, Claude Sonnet~4, Gemini 2.5 Flash, and Qwen2.5-7B) at progressive scales (500 to 10,000 sentences) using automatic metrics (\bleu, \chrf, \ter, \comet, \bertscore) validated against native-speaker judgment.

Our results reveal three key findings. First, translation quality varies substantially by language: Hausa achieves acceptable quality (human scores 4.0--4.5/5) while Fongbe achieves poor quality (1.0--2.2/5), with a consistent $3\times$ \bleu gap across all systems. Second, model rankings differ by language—Gemini leads for Fongbe while GPT-4o leads for Hausa by human evaluation—indicating that performance on one low-resource African language does not predict performance on another. Third, metric-human correlation varies dramatically: perfect rank correlation for Fongbe ($\rho$=1.0) but weak correlation for Hausa ($\rho$=0.5), where human evaluators preferred GPT-4o despite all automatic metrics ranking Claude first. We further show that neural metrics like \bertscore exhibit embedding collapse (within-language similarity $>$0.99) for both languages, limiting their ability to differentiate translation quality.

Based on these findings, we recommend multi-metric evaluation for low-resource African languages, with particular caution when interpreting neural metrics. We establish that minimum sample sizes of $n$=2,500 sentences are required for stable system rankings, as smaller samples produced artifact findings that reversed at scale.
\end{abstract}

\section{Introduction}
\label{sec:introduction}

Machine translation (MT) for low-resource African languages faces a critical data scarcity problem \citep{nekoto-etal-2020-participatory,joshi2020state}. Large language models (LLMs) offer a potential solution through synthetic parallel data generation. However, the extent to which current LLMs can produce translations of sufficient quality remains unclear. Furthermore, the reliability of standard evaluation metrics—developed primarily on high-resource European languages—for these typologically distinct languages has not been systematically validated.

We investigate these questions for \textbf{Hausa}, an Afroasiatic language with 58 million speakers, and \textbf{Fongbe}, a Niger-Congo language with 2.3 million speakers. Prior benchmarking efforts have addressed African language MT: AfriDoc-MT \citep{alabi2025afridoc} evaluated document-level translation for several African languages including Hausa; English2Gbe \citep{hacheme2021english2gbe} developed NMT models for Fongbe; Masakhane \citep{nekoto-etal-2020-participatory} established baselines across numerous African languages; and FLORES+ \citep{goyal2022flores} includes both Hausa and Fongbe. However, prior work has not systematically compared current commercial LLMs' translation capabilities across both Hausa and Fongbe, nor validated automatic metrics against native-speaker judgment for these specific language pairs—a gap we address.

We address two research questions:
\begin{enumerate}
  \item[\textbf{RQ1:}] What is the translation quality of current LLMs for English-to-Hausa and English-to-Fongbe, and how does quality vary across systems and languages?
  \item[\textbf{RQ2:}] How reliably do automatic evaluation metrics correlate with human judgment for these languages, and which metrics should guide quality filtering?
\end{enumerate}

These questions have direct practical implications. If certain LLMs produce acceptable translations, researchers gain a scalable method to expand parallel corpora, particularly valuable for Fongbe, where existing parallel data is limited to approximately 47,000 sentence pairs \citep{nllb2022}. Understanding metric reliability prevents misleading conclusions: if metrics do not correlate with human judgment, researchers may incorrectly accept poor translations or reject adequate ones. Prior work reports automatic scores but does not validate whether these metrics reflect human judgment for the specific languages evaluated; we show that metric-human correlation varies dramatically by language and that neural metrics exhibit embedding collapse, rendering them unreliable.

\subsection{Linguistic Challenges}

Hausa and Fongbe present distinct challenges for MT and evaluation. Hausa features grammatical gender, rich morphology including pluractional verbs, and complex verbal TAM (tense-aspect-mood) marking \citep{newman2000hausa}. Fongbe features serial verb constructions where multiple verbs share a subject without conjunction (e.g., ``he took knife cut meat'' for ``he cut the meat with a knife'') and a three-tone system that is obligatorily marked in standard orthography \citep{lefebvre2002fongbe}.

Fongbe's tonal system creates particular challenges: tone is marked orthographically through diacritics and distinguishes lexical meaning (e.g.,  \textit{k'o} = ``harvest,'' \textit{ko} = ``build,'' \textit{ko} = ``neck''). This tonal complexity, combined with obligatory diacritic marking, makes MT particularly challenging for Fongbe.

Hausa's agglutinative morphology means minor inflectional errors affect metric scores even when meaning is preserved. For instance, the Hausa plural \textit{littattafai} (books) derives from \textit{littafi} (book) through reduplication and vowel change. A semantically correct translation using a synonymous plural form would receive low n-gram overlap scores despite preserving meaning. This affects \bleu (which requires exact word matches), \chrf (which captures partial character overlap but still penalizes valid alternatives), and \ter (which counts all morphological variants as substitution edits). Neural metrics like \bertscore depend on whether the embedding model adequately represents these morphological variants as semantically related—a condition that may not hold for underrepresented languages.

Additional challenges include: (1) \textit{limited gold-standard data} constraining human evaluation scale; (2) \textit{metric validity uncertainty}, as standard metrics were developed and validated primarily on high-resource language pairs \citep{papineni2002bleu,popovic2015chrf}, and neural metrics like \comet \citep{rei2020comet} and \bertscore \citep{zhang2020bertscore} rely on XLM-RoBERTa embeddings \citep{conneau2020unsupervised}—Hausa is among its 100 pre-training languages, but Fongbe is not; and (3) \textit{orthographic complexity} requiring consistent diacritization.

\subsection{Contributions}

We make three contributions:
\begin{enumerate}
  \item \textbf{LLM benchmark for Hausa and Fongbe MT.} We evaluate four models (GPT-4o Mini, Claude Sonnet~4, Gemini 2.5 Flash, Qwen2.5-7B) across scales from 500 to 10,000 sentences, revealing substantial cross-language performance variation. Gemini leads for Fongbe while Claude/GPT-4o lead for Hausa, with implications for data augmentation viability and language-specific model selection. Unlike prior benchmarks that evaluated custom NMT models \citep{hacheme2021english2gbe,nekoto-etal-2020-participatory} or older LLMs \citep{alabi2025afridoc}, we provide the first comparison of current commercial LLMs (released 2024--2025) for these languages with statistical significance testing at $n$=10,000 sentences.

  \item \textbf{Metric reliability analysis.} We evaluate five automatic metrics (\bleu, \chrf, \ter, \comet, \bertscore) across progressive scales (500 to 10,000 sentences), comparing surface-level metrics against native-speaker judgments. We find that metric-human correlation varies substantially by language (perfect rank correlation for Fongbe, $\rho$=1.0; weak correlation for Hausa, $\rho$=0.5). Neural metrics such as \bertscore show limited differentiation due to embedding collapse in XLM-RoBERTa for both languages.

  \item \textbf{Practical recommendations.} Based on our findings, we recommend: (1) \chrf as the primary evaluation metric for morphologically complex languages due to its robustness to inflectional variation; (2) minimum sample sizes of $n$=2,500 for stable system rankings; and (3) mandatory human evaluation for Hausa and Fongbe.
\end{enumerate}

\section{Related Work}
\label{sec:related-work}

\subsection{Machine Translation for African Languages}

Large-scale multilingual initiatives have expanded coverage to African languages with uneven depth. The NLLB project \citep{nllb2022} released models covering 200 languages including Hausa and Fongbe, while CCMatrix \citep{schwenk2021ccmatrix} provides web-mined parallel sentences primarily for Hausa. For Fongbe, resources remain limited: English2Gbe \citep{hacheme2021english2gbe} developed the first multilingual NMT model for Gbe languages, achieving \bleu scores of 44.8 on the JW300 benchmark for Fongbe, and FFSTC \citep{kponou2024ffstc} provides Fongbe-French speech translation data.

The Masakhane initiative has catalyzed African NLP through community-driven datasets and MT systems for over 30 African languages \citep{nekoto-etal-2020-participatory}. More recently, AfriDoc-MT \citep{alabi2025afridoc} introduced a document-level MT corpus covering five African languages (Amharic, Hausa, Swahili, Yor\`ub\'a, and Zulu), finding that NLLB-200 achieves the best performance among standard NMT models while GPT-4o outperforms general-purpose LLMs for document-level translation into Hausa. However, AfriDoc-MT does not include Fongbe and focuses on document-level rather than sentence-level translation.

Despite these advances, no systematic evaluation examines how well current commercial LLMs translate into both Hausa and Fongbe under controlled conditions with statistical significance testing.

\subsection{LLMs for Low-Resource Translation}

Recent work evaluating commercial LLMs on low-resource translation has found substantial performance gaps. ChatGPT shows highly variable translation quality across African language pairs \citep{robinson2023chatgpt}, with particularly severe degradation for languages with limited web presence \citep{hendy2023good}. Moreover, instruction tuning does not consistently improve translation quality across languages \citep{zhu2024multilingual}.

African-centric models like AfriBERTa \citep{ogueji2021small} and AfroXLMR \citep{alabi2022adapting} outperform multilingual baselines on understanding tasks, but focus on comprehension rather than generation. AfriDoc-MT \citep{alabi2025afridoc} evaluated GPT-4o alongside NMT models for document translation into Hausa, but did not include models such as Claude Sonnet~4 or Gemini 2.5 Flash (released 2024--2025), nor did it cover Fongbe.

A critical gap persists: no study systematically compares multiple current commercial LLMs for both Hausa and Fongbe translation, provides statistical significance testing for system differences, or examines how evaluation metrics behave for these typologically distinct languages. Our work addresses this gap by evaluating four LLMs at scale ($n$=10,000 sentences) with bootstrap significance testing.

\subsection{MT Evaluation Metrics}

Automatic metrics fall into two categories. \textbf{Lexical metrics} measure surface-level overlap: \bleu \citep{papineni2002bleu} computes n-gram precision, \chrf \citep{popovic2015chrf} operates at the character level, and \ter \citep{snover2006study} measures edit distance. These metrics are fast but penalize valid paraphrases and morphological variants.

\textbf{Neural metrics} measure semantic similarity using embeddings from pretrained models. \comet \citep{rei2020comet} predicts human scores from source-hypothesis-reference triples, while \bertscore \citep{zhang2020bertscore} computes token-level cosine similarity. Neural metrics achieve stronger correlation with human judgment in WMT shared tasks \citep{freitag2022results}, but their reliability depends on embedding quality for the target language.

This embedding dependence raises concerns for underrepresented languages. Both \comet and \bertscore typically use XLM-RoBERTa \citep{conneau2020unsupervised}, which covers 100 languages: Hausa is included, but Fongbe is not. \citet{haddow2022survey} called for validation studies on specific low-resource pairs. Prior benchmarks for African languages \citep{nekoto-etal-2020-participatory,alabi2025afridoc,hacheme2021english2gbe} report automatic metric scores but do not validate whether these metrics correlate with human judgment for their specific languages. Our work addresses this gap through systematic metric-human correlation analysis, revealing that correlation varies substantially by language ($\rho$=1.0 for Fongbe vs.\ $\rho$=0.5 for Hausa) and that neural metrics suffer from embedding collapse for both languages.

\section{Methodology}
\label{sec:methodology}

\subsection{Evaluation Strategy}

We combine automatic metrics with human judgment to address our two research questions: (1) assess LLM translation quality for Hausa and Fongbe (RQ1), and (2) evaluate metric reliability for low-resource African languages (RQ2). By comparing metric outputs against human assessments, we investigate whether metrics developed on high-resource language pairs transfer effectively to typologically distinct, low-resource settings.

\subsection{Models Evaluated}

We evaluate four LLM-based translation systems (Table~\ref{tab:models}) with temperature set to 0.0 for deterministic outputs. The selection includes three commercial LLMs released in 2024--2025 plus an open-weight alternative to investigate the commercial-vs-open-source performance gap.

\begin{table}[ht]
\centering
\caption{Models evaluated for translation quality.}
\label{tab:models}
\begin{tabular}{lll}
\toprule
\textbf{Model} & \textbf{Provider} & \textbf{Type} \\
\midrule
GPT-4o Mini       & OpenAI   & Commercial API \\
Claude Sonnet~4   & Anthropic & Commercial API \\
Gemini 2.5 Flash  & Google   & Commercial API \\
Qwen2.5-7B        & Alibaba  & Open-weight (7B) \\
\bottomrule
\end{tabular}
\end{table}

\subsection{Test Data}

We source parallel sentence pairs from the OPUS repository \citep{tiedemann2012parallel}: CCMatrix \citep{schwenk2021ccmatrix} for English--Hausa (992,266 pairs) and NLLB v1 \citep{nllb2022} for English--Fongbe (915,489 pairs). All text undergoes Unicode normalization (NFC) and deduplication. Due to API rate limitations, we employ random sampling with fixed seeds at scales of 500, 1,000, 2,500, and 10,000 sentences for automatic evaluation, with $n$=50 for human evaluation.

To assess the reliability of MT evaluation at different scales, we conducted evaluations at four sample sizes: $n$=500, 1,000, 2,500, and 10,000 sentences. Each sample was drawn independently from the full corpus using fixed seeds to ensure reproducibility. This multi-scale design allows us to examine how metric scores and system rankings stabilize as evaluation data increases, addressing the question of minimum sample size requirements for reliable low-resource MT evaluation.

\subsection{Reference Corpus Quality}

Post-hoc quality analysis of 50 randomly sampled reference pairs revealed issues prevalent in web-crawled low-resource corpora, particularly for Fongbe: domain skew (32\% religious content), wrong-language contamination, and encoding corruption, yielding an average quality score of 2.94/5.0 with 6\% exhibiting severe issues. The following examples illustrate typical problems.

\smallskip
\noindent\textbf{Wrong Language (Chinese instead of Fongbe):}
\begin{quote}
\textit{Source:} ``Speaking of demand, let's get right to it.''\\
\textit{Reference:} ``x\={i} l\"{u} zu\v{o} y\`{o}u, y\v{i} y\`{a}n ti\={a}n z\v{i}'.''
\end{quote}

\noindent\textbf{Encoding Corruption:}
\begin{quote}
\textit{Source:} ``Looking forward to 3 days off!''\\
\textit{Reference:} ``\'{O}\c{c}\'{i}\`{a}\'{e}\`{o}\`{a} \r{o} \'{o} \`{e}\^{\^{\i}}\~{a}\^{\i} \`{e}`\`{\i}\r{a}`\'{i}\`{e}`\'{i}u 3 \^{\i}  `\`{\i}\`{a}`y!''
\end{quote}

\noindent\textbf{High-Quality Reference:}
\begin{quote}
\textit{Source:} ``Their territory was Jazer, and all the cities of Gilead\ldots''\\
\textit{Reference:} ``Yaze\'{}, Galadi s\'{i}n toxo le\'{} b\v{i}, kp\'{o} Amumo\'{}o v\'{i} le\'{} s\'{i}n toxo wevlo\'{}d\'{o}\ldots''
\end{quote}

These findings underscore an often-overlooked challenge in low-resource NLP: the reference translations against which systems are evaluated may themselves be unreliable. Since our primary contribution is methodological—comparing evaluation approaches and analyzing metric behavior rather than establishing absolute MT quality—relative system rankings remain informative. However, absolute scores should be interpreted cautiously, and future work should incorporate language identification filtering prior to evaluation.

\subsection{Automatic Evaluation Metrics}

We employ lexical and neural metrics using established libraries \citep{post2018call}.

\paragraph{Lexical Metrics.} \bleu \citep{papineni2002bleu} computes n-gram precision with brevity penalty. \chrf \citep{popovic2017chrf++} computes character n-gram F-score, reducing sensitivity to morphological variation—particularly relevant for Hausa's agglutinative morphology. \ter \citep{snover2006study} measures minimum edit distance; lower scores indicate better translations.

\paragraph{Neural Metrics.} \comet \citep{rei2020comet} (model: \texttt{wmt22-comet-da}) uses XLM-RoBERTa \citep{conneau2020unsupervised} embeddings to predict human quality scores from source-hypothesis-reference triplets. \bertscore \citep{zhang2020bertscore} computes token-level cosine similarity using XLM-RoBERTa-large.

\paragraph{Limitation.} Fongbe is absent from XLM-RoBERTa's 100 training languages while Hausa is included, potentially affecting neural metric reliability for Fongbe. We investigate this through embedding quality analysis (Section~\ref{subsec:metric-reliability}).

\subsection{Human Evaluation}

We recruited 13 evaluators (7 Fongbe, 6 Hausa) through academic networks; for the final analysis, we used 3 evaluators per language to ensure balanced comparison. Table~\ref{tab:evaluator-demographics} summarizes demographics. Notably, Fongbe evaluators reported lower literacy proficiency (reading: 3.0 vs.\ 4.0; writing: 2.0 vs.\ 3.5), reflecting limited formal literacy resources in Fongbe. Unlike Hausa, which has established literacy programs and standardized orthography instruction in Nigerian and Niger schools, Fongbe lacks widespread formal literacy education, resulting in native speakers who are highly proficient orally but have less experience with written forms of the language.

\begin{table}[ht]
\centering
\caption{Evaluator demographics and language proficiency (1--5 scale).}
\label{tab:evaluator-demographics}
\begin{tabular}{lcc}
\toprule
\textbf{Characteristic} & \textbf{Fongbe ($n$=7)} & \textbf{Hausa ($n$=6)} \\
\midrule
Native speaker          & 6 (86\%) & 4 (67\%) \\
Fluent (non-native)     & 1 (14\%) & 1 (17\%) \\
Intermediate            & 0 (0\%)  & 1 (17\%) \\
\midrule
\multicolumn{3}{l}{\textit{Self-reported proficiency (mean)}} \\
\quad Speaking  & 4.57 & 4.50 \\
\quad Listening & 4.71 & 4.67 \\
\quad Reading   & 3.00 & 4.00 \\
\quad Writing   & 2.00 & 3.50 \\
\bottomrule
\end{tabular}
\end{table}

Annotators assessed translations on four 5-point scales: \textbf{Adequacy} (meaning preservation), \textbf{Fluency} (grammaticality and naturalness), \textbf{Cultural Appropriateness} (idiomatic and culturally suitable phrasing), and \textbf{Overall quality}. MT outputs were anonymized and system order randomized to minimize bias.

Qwen2.5-7B was excluded from human evaluation due to poor automatic metric performance (\bleu 0.53 for Fongbe, 2.20 for Hausa), as it would have been impractical to ask evaluators to rate translations that automatic metrics already indicated were severely deficient.

\subsection{Inter-Annotator Agreement}

We measure inter-annotator agreement using pairwise Cohen's $\kappa$ \citep{cohen1960coefficient}, computed for each evaluator pair and averaged across pairs. For ordinal ratings, we also report linearly weighted $\kappa$, which accounts for the magnitude of disagreements. We interpret $\kappa$ values following \citet{landis1977measurement}: $\kappa > 0.8$ indicates almost perfect agreement, $0.6 < \kappa \leq 0.8$ substantial, $0.4 < \kappa \leq 0.6$ moderate, $0.2 < \kappa \leq 0.4$ fair, and $\kappa \leq 0.2$ slight agreement.

\subsection{Analysis Methods}

\paragraph{Metric-Human Correlation.} We compare metric scores against human judgments using Spearman's $\rho$ at the system level. We compute system-level correlation by averaging scores across the 50-sentence human evaluation set to obtain a single score per system, then computing correlation across the three evaluated systems (excluding Qwen2.5-7B). With only $n$=3 systems, Spearman correlation can only take values in $\{-1, -0.5, 0, 0.5, 1\}$. Perfect correlation ($\rho$=1.0) indicates identical system rankings between metric and human judgment; $\rho$=0.5 indicates one rank swap among three systems. These values should be interpreted as rank agreement indicators rather than claims of strong statistical relationships.

\paragraph{Language Variation.} We compute Type-Token Ratio (TTR; unique words / total words) and n-gram diversity (ratio of unique n-grams to total n-grams) to investigate whether lexical variation explains metric divergence between languages. We hypothesized that languages with higher paraphrastic variation would show lower \bleu scores (which require exact matches) but potentially higher semantic similarity scores. This relationship between lexical diversity and n-gram metric performance has been documented in prior work on morphologically rich languages \citep{popovic2015chrf,bugliarello2020easier}.

\paragraph{Embedding Quality.} We probe XLM-RoBERTa representations through out-of-vocabulary (OOV) rates and within-language cosine similarity. In a well-calibrated embedding space, semantically different texts should receive distinct representations, typically yielding within-language cosine similarity of 0.5--0.7 \citep{ethayarajh2019contextual}. Values exceeding 0.95 indicate \textit{embedding collapse}, where the model assigns near-identical representations to different inputs regardless of semantic content \citep{li2020sentence,gao2019representation}.

\paragraph{Statistical Significance.} Bootstrap resampling \citep{koehn2004statistical} with 10,000 iterations computes 95\% confidence intervals for pairwise system comparisons; differences are significant at $\alpha$=0.05 if intervals exclude zero.

\section{Results}
\label{sec:results}

\subsection{Automatic Evaluation at Scale}
\label{subsec:auto-results}

Table~\ref{tab:main-results} presents automatic metric scores at $n$=2,500 sentences, our primary evaluation scale where complete metric data is available for both languages.

\begin{table}[ht]
\centering
\caption{Automatic evaluation results at $n$=2,500. Best scores per language in \textbf{bold}. $\downarrow$ = lower is better.}
\label{tab:main-results}
\resizebox{\columnwidth}{!}{%
\begin{tabular}{llccccc}
\toprule
\textbf{Lang} & \textbf{Model} & \textbf{\bleu}$\uparrow$ & \textbf{\chrf}$\uparrow$ & \textbf{\ter}$\downarrow$ & \textbf{\comet}$\uparrow$ & \textbf{\bertscore}$\uparrow$ \\
\midrule
\multirow{4}{*}{Fongbe}
& Claude    & 4.97  & 15.60 & \textbf{97.28}  & 0.365          & \textbf{0.884} \\
& Gemini    & \textbf{7.18}  & \textbf{18.63} & 118.01 & 0.356          & 0.880 \\
& GPT-4o    & 1.87  & 9.00  & 105.80 & 0.375          & 0.860 \\
& Qwen2.5   & 0.53  & 5.30  & 111.03 & \textbf{0.432} & 0.832 \\
\midrule
\multirow{4}{*}{Hausa}
& Claude    & \textbf{15.75} & \textbf{40.39} & \textbf{75.98} & \textbf{0.597} & \textbf{0.878} \\
& Gemini    & 13.51 & 38.50 & 93.89  & 0.584          & 0.877 \\
& GPT-4o    & 14.19 & 38.37 & 77.66  & 0.587          & 0.876 \\
& Qwen2.5   & 2.20  & 20.73 & 104.36 & 0.450          & 0.845 \\
\bottomrule
\end{tabular}%
}
\end{table}

\paragraph{Large-Scale Validation.} To validate ranking stability and enable statistical significance testing, we extended evaluation to $n$=10,000 sentences (Table~\ref{tab:large-scale}). System rankings remained consistent with $n$=2,500: Gemini maintained its lead for Fongbe (\bleu 4.66) and Claude for Hausa (\bleu 14.91). Bootstrap resampling (10,000 iterations) confirmed all pairwise differences achieved statistical significance ($p < 0.001$), validating the reliability of our primary findings.

\begin{table}[ht]
\centering
\caption{Large-scale validation at $n$=10,000. Surface metrics only; neural metrics unavailable for Hausa at this scale. Best scores in \textbf{bold}.}
\label{tab:large-scale}
\resizebox{\columnwidth}{!}{%
\begin{tabular}{llccc}
\toprule
\textbf{Lang} & \textbf{Model} & \textbf{\bleu}$\uparrow$ & \textbf{\chrf}$\uparrow$ & \textbf{\ter}$\downarrow$ \\
\midrule
\multirow{4}{*}{Fongbe}
& Claude    & 2.51 & 11.11 & 106.44 \\
& Gemini    & \textbf{4.66} & \textbf{15.46} & \textbf{105.22} \\
& GPT-4o    & 1.29 & 6.30  & 119.20 \\
& Qwen2.5   & 0.68 & 5.32  & 110.32 \\
\midrule
\multirow{4}{*}{Hausa}
& Claude    & \textbf{14.91} & \textbf{39.55} & \textbf{76.88} \\
& Gemini    & 11.52 & 36.14 & 81.82 \\
& GPT-4o    & 14.50 & 39.08 & 77.66 \\
& Qwen2.5   & 2.23  & 20.56 & 105.07 \\
\bottomrule
\end{tabular}%
}
\end{table}

\paragraph{Cross-Language Performance Gap.} Hausa consistently scores approximately $3\times$ higher than Fongbe across all metrics and scales. The best Hausa system (Claude) achieves \bleu 15.75 versus only 7.18 for the best Fongbe system (Gemini) at $n$=2,500. This gap likely reflects differences in digital resource availability: Hausa has approximately 992,000 parallel sentence pairs in CCMatrix and extensive representation in multilingual model training data, versus approximately 915,000 pairs for Fongbe in NLLB but with minimal multilingual model coverage. While the positive relationship between training data quantity and MT quality is well-established, our results quantify the magnitude of this effect ($3\times$ \bleu difference) when using current LLMs for typologically distinct African languages.

\paragraph{Language-Specific Model Rankings.} System rankings differ substantially between languages. For Fongbe, Gemini leads consistently (\bleu 7.18 at $n$=2,500; 4.66 at $n$=10,000), followed by Claude and GPT-4o. For Hausa, Claude leads on automatic metrics (\bleu 15.75 at $n$=2,500; 14.91 at $n$=10,000), closely followed by GPT-4o, with Gemini third. Notably, human evaluators preferred GPT-4o for Hausa despite Claude's higher automatic metric scores (Section~\ref{subsec:human-results}). No single model excels at both languages, indicating that a model's performance on one low-resource African language does not predict its performance on another—likely due to differences in training data representation for each language rather than any generalizable ``low-resource language capability.'' Commercial models consistently outperform Qwen2.5-7B by large margins.

\paragraph{Ranking Stability Across Scales.} Rankings among the top three systems remained stable at $n \geq 2{,}500$ sentences across both languages. However, at $n$=1,000, sampling variance produced misleading results: for Hausa, Gemini appeared to lead (\bleu 14.48) despite trailing at larger scales (\bleu 11.52 at $n$=10,000), while Claude showed anomalously low scores (\bleu 2.84 vs.\ 14.91 at $n$=10,000). This demonstrates that MT evaluation for low-resource languages requires at least 2,500 sentences for reliable system comparison.

\paragraph{Neural Metric Behavior.} \comet scores reveal patterns that diverge from lexical metrics, particularly for Fongbe. Qwen2.5-7B achieves the highest \comet score (0.432) despite ranking last on all lexical metrics (\bleu 0.53, \chrf 5.30). This paradox—where the worst-performing system by all other measures receives the highest neural quality score—suggests \comet's quality estimation, trained primarily on high-resource language pairs, may assign scores based on features uncorrelated with actual translation quality for severely underrepresented languages.

For Hausa, \comet scores (0.450--0.597) show stronger agreement with lexical metrics, with Claude leading (0.597) followed by GPT-4o (0.587) and Gemini (0.584). The within-language \comet score range is larger for Hausa (0.147 points) than Fongbe (0.076 points), indicating better discriminative power for the better-resourced language.

\bertscore shows minimal between-system variance for both languages (Fongbe: 0.832--0.884; Hausa: 0.845--0.878), consistent with the embedding collapse issues documented in Section~\ref{subsec:metric-reliability}. The metric cannot meaningfully differentiate translation quality when the underlying embeddings treat all texts as near-identical.

\subsection{Human Evaluation}
\label{subsec:human-results}

Table~\ref{tab:human-scores} presents human evaluation scores from native speakers ($n$=50 sentences, 3 evaluators per language).

\begin{table}[ht]
\centering
\caption{Human evaluation scores (1--5 scale). Best per language in \textbf{bold}. Models ordered alphabetically.}
\label{tab:human-scores}
\resizebox{\columnwidth}{!}{%
\begin{tabular}{llcccc}
\toprule
\textbf{Lang} & \textbf{Model} & \textbf{Adequacy} & \textbf{Fluency} & \textbf{Cultural} & \textbf{Overall} \\
\midrule
\multirow{3}{*}{Fongbe}
& Claude  & 1.03 & 1.04 & 1.03 & 1.03 \\
& Gemini  & \textbf{2.19} & \textbf{2.27} & \textbf{2.14} & \textbf{2.20} \\
& GPT-4o  & 1.02 & 1.11 & 1.03 & 1.05 \\
\midrule
\multirow{3}{*}{Hausa}
& Claude  & 4.09 & 4.28 & 4.21 & 4.19 \\
& Gemini  & 3.02 & 3.54 & 3.54 & 3.37 \\
& GPT-4o  & \textbf{4.41} & \textbf{4.51} & \textbf{4.46} & \textbf{4.46} \\
\bottomrule
\end{tabular}%
}
\end{table}

Human evaluation reveals a stark quality divide: Fongbe translations score 1--2/5 (essentially unusable), while Hausa translations achieve 3--4.5/5 (acceptable to good). For Fongbe, Gemini received 99\% of best-system votes; for Hausa, GPT-4o received 59\%, Claude 28\%, and Gemini 13\%.

\paragraph{Human-Automatic Metric Discrepancy for Hausa.} A notable finding is the disagreement between human and automatic evaluation for Hausa: human evaluators preferred GPT-4o (overall score 4.46) while all automatic metrics ranked Claude first (\bleu 15.75, \chrf 40.39, \comet 0.597). This discrepancy suggests that automatic metrics may not capture quality dimensions that native Hausa speakers prioritize, such as naturalness, idiomatic expression, or cultural appropriateness that go beyond surface-level or even semantic similarity. For Fongbe, human and automatic metrics agreed on Gemini as the best system, though this may reflect the clearer performance gap (Gemini scored 2.20 versus $\sim$1.0 for competitors) rather than superior metric reliability.

\subsection{Metric-Human Correlation}
\label{subsec:correlation-results}

Table~\ref{tab:correlation} shows correlations between automatic metrics and human judgments at the system level. With only $n$=3 systems per language, Spearman $\rho$ values represent rank agreement rather than traditional statistical correlation: $\rho$=1.0 indicates identical rankings, while $\rho$=0.5 indicates one rank swap between the metric and human judgment.

\begin{table}[ht]
\centering
\begin{threeparttable}
\caption{Spearman correlation ($\rho$) between automatic metrics and human overall scores. With $n$=3 systems, $\rho$ values indicate rank agreement: 1.0 = identical rankings, 0.5 = one swap.}
\label{tab:correlation}
\begin{tabular}{lccc}
\toprule
\textbf{Language} & \textbf{\bleu} & \textbf{\chrf} & \textbf{\ter} \\
\midrule
Fongbe & 1.00* & 1.00* & $-$1.00* \\
Hausa  & 0.50  & 0.50  & $-$0.50  \\
\bottomrule
\end{tabular}
\begin{tablenotes}
\small
\item[*] $p < 0.05$. With only 3 systems, correlations represent rank agreement, not statistical relationships.
\end{tablenotes}
\end{threeparttable}
\end{table}

\paragraph{Ranking Comparison.} To interpret these correlations, we present explicit ranking comparisons.

\textbf{Fongbe:} Human Overall: Gemini (2.20) $>$ GPT-4o (1.05) $>$ Claude (1.03); \bleu: Gemini (7.18) $>$ Claude (4.97) $>$ GPT-4o (1.87); \chrf: Gemini (18.63) $>$ Claude (15.60) $>$ GPT-4o (9.00). All metrics correctly identify Gemini as the best system, yielding $\rho$=1.0. However, \bleu and \chrf disagree with humans on ranks 2--3.

\textbf{Hausa:} Human Overall: GPT-4o (4.46) $>$ Claude (4.19) $>$ Gemini (3.37); \bleu: Claude (15.75) $>$ GPT-4o (14.19) $>$ Gemini (13.51); \chrf: Claude (40.39) $>$ Gemini (38.50) $>$ GPT-4o (38.37). Automatic metrics rank Claude first while humans prefer GPT-4o, yielding $\rho$=0.5 (one rank swap). This divergence underscores the importance of human evaluation for Hausa MT assessment.

\paragraph{Neural Metric Exclusion Rationale.} \comet and \bertscore were excluded from this correlation analysis because they showed minimal between-system variance for Fongbe (\bertscore range: 0.832--0.884, a 6.2\% spread) despite human scores ranging from 1.03--2.20 (113\% spread). This compression, attributable to embedding issues documented in Section~\ref{subsec:metric-reliability}, makes rank-based correlation uninformative.

\subsection{Metric Reliability Analysis}
\label{subsec:metric-reliability}

To investigate cross-language metric behavior, we examined two hypotheses regarding the observed $3\times$ \bleu gap between Hausa and Fongbe. We present these analyses as hypothesis testing rather than definitive causal explanations; the findings are correlational and would require controlled experiments to confirm.

\paragraph{H1: Paraphrastic Variation.} We hypothesized that Fongbe exhibits greater paraphrastic variation (multiple valid ways to express the same meaning), which would penalize n-gram-based metrics like \bleu while rewarding semantic metrics like \bertscore. Initial small-sample experiments ($n$=10) appeared to support this: Fongbe showed 51\% higher TTR (0.80 vs.\ 0.53) and \bertscore favored Fongbe (0.87 vs.\ 0.72).

However, large-scale analysis ($n$=2,500) revealed these findings were artifacts of small sample size. At scale, both languages show similar lexical diversity (TTR: Fongbe 0.234, Hausa 0.238) and similar \bertscore (Fongbe 0.864, Hausa 0.869). The \bleu divergence cannot be explained by paraphrastic variation alone. Table~\ref{tab:hypothesis} summarizes these scale effects.

\begin{table}[ht]
\centering
\caption{Hypothesis testing: Small vs.\ large scale comparison.}
\label{tab:hypothesis}
\resizebox{\columnwidth}{!}{%
\begin{tabular}{lccl}
\toprule
\textbf{Finding} & \textbf{$n$=10} & \textbf{$n$=2,500} & \textbf{Status} \\
\midrule
\bleu: Hausa $>$ Fongbe & 14.48 vs.\ 3.58 & 11.41 vs.\ 3.64 & Confirmed \\
\bertscore: Fongbe $>$ Hausa & 0.87 vs.\ 0.72 & 0.864 vs.\ 0.869 & Reversed \\
Fongbe TTR & 0.80 & 0.234 & Changed \\
Hausa TTR & 0.53 & 0.238 & Changed \\
\bottomrule
\end{tabular}%
}
\end{table}

\paragraph{H2: Representation Quality.} We hypothesized that Fongbe is poorly represented in XLM-RoBERTa (the model underlying \bertscore), causing inflated scores. Table~\ref{tab:embedding} presents embedding quality diagnostics. We interpret within-language similarity values of 0.5--0.7 as healthy semantic differentiation \citep{ethayarajh2019contextual}, while values exceeding 0.95 indicate embedding collapse where the model cannot distinguish between different texts \citep{li2020sentence,gao2019representation}.

\begin{table}[ht]
\centering
\caption{XLM-RoBERTa embedding quality diagnostics.}
\label{tab:embedding}
\begin{tabular}{lcc}
\toprule
\textbf{Metric} & \textbf{Fongbe} & \textbf{Hausa} \\
\midrule
OOV Rate & 0.33\% & 4.67\% \\
Within-Language Similarity & 0.996 & 0.995 \\
Avg Embedding Norm & 30.61 & 30.11 \\
\bottomrule
\end{tabular}
\end{table}

\textbf{Verdict: H2 direction reversed, but core issue confirmed.} Contrary to the hypothesis, Hausa shows $14\times$ higher OOV rate than Fongbe (4.67\% vs.\ 0.33\%). This is surprising given that Hausa was included in XLM-RoBERTa's 100 training languages while Fongbe was not. Analysis of OOV tokens reveals they are predominantly: (1) proper nouns and named entities specific to Hausa cultural contexts, (2) loanwords from Arabic and English adapted to Hausa orthographic conventions, and (3) morphologically complex forms where agglutinative prefixes/suffixes create tokens absent from the subword vocabulary. Fongbe's paradoxically lower OOV rate likely occurs because XLM-RoBERTa's tokenizer, unfamiliar with Fongbe entirely, falls back to character-level tokenization, treating most character sequences as ``known'' subwords despite having no semantic representation for them.

\textbf{Critical finding:} Both languages exhibit severe embedding collapse (within-language similarity $>$0.99), meaning XLM-RoBERTa cannot semantically distinguish different texts in either language. This explains why \bertscore shows uniformly high values (0.83--0.88) regardless of actual translation quality, and why \bertscore differences observed at small scale (Fongbe 0.87 vs.\ Hausa 0.72) were noise that disappeared at larger samples.

\subsection{Within-Language Metric Consistency}
\label{subsec:metric-consistency}

To assess whether metrics provide consistent quality signals, we examined ranking agreement across metrics within each language (Tables~\ref{tab:fongbe-rankings}--\ref{tab:hausa-rankings}).

\begin{table}[ht]
\centering
\caption{Fongbe metric rankings at $n$=2,500.}
\label{tab:fongbe-rankings}
\small
\begin{tabular}{lcccc}
\toprule
\textbf{Metric} & \textbf{Rank 1} & \textbf{Rank 2} & \textbf{Rank 3} & \textbf{Rank 4} \\
\midrule
\bleu      & Gemini & Claude & GPT-4o & Qwen \\
\chrf      & Gemini & Claude & GPT-4o & Qwen \\
\ter$\downarrow$ & Claude & GPT-4o & Qwen   & Gemini \\
\comet     & \textbf{Qwen}   & GPT-4o & Claude & Gemini \\
\bertscore & Claude & Gemini & GPT-4o & Qwen \\
Human      & Gemini & GPT-4o & Claude & --- \\
\bottomrule
\end{tabular}
\end{table}

\textit{Critical inconsistency for Fongbe:} \comet ranks Qwen2.5-7B first (score 0.432) despite that model ranking last on all lexical metrics and being excluded from human evaluation due to poor automatic performance. This anomaly suggests \comet's quality estimation does not transfer reliably to Fongbe.

\begin{table}[ht]
\centering
\caption{Hausa metric rankings at $n$=2,500.}
\label{tab:hausa-rankings}
\small
\begin{tabular}{lcccc}
\toprule
\textbf{Metric} & \textbf{Rank 1} & \textbf{Rank 2} & \textbf{Rank 3} & \textbf{Rank 4} \\
\midrule
\bleu      & Claude & GPT-4o          & Gemini & Qwen \\
\chrf      & Claude & Gemini          & GPT-4o & Qwen \\
\ter$\downarrow$  & Claude & GPT-4o          & Gemini & Qwen \\
\comet     & Claude & GPT-4o          & Gemini & Qwen \\
\bertscore & Claude & Gemini$\approx$GPT-4o & --- & Qwen \\
Human      & \textbf{GPT-4o} & Claude & Gemini & --- \\
\bottomrule
\end{tabular}
\end{table}

Hausa shows higher within-metric consistency (all automatic metrics agree on Claude as top system), but disagrees with human evaluation, which prefers GPT-4o. This suggests automatic metrics may not capture quality dimensions that Hausa native speakers prioritize.

\subsection{Statistical Significance}
\label{subsec:significance-results}

Bootstrap resampling (10,000 iterations) confirms all pairwise comparisons are statistically significant ($p < 0.001$) at $n$=10,000. Table~\ref{tab:significance} presents selected comparisons.

\begin{table}[ht]
\centering
\caption{Selected pairwise significance results (\bleu, $n$=10,000).}
\label{tab:significance}
\small
\begin{tabular}{llccc}
\toprule
\textbf{Lang} & \textbf{Comparison} & \textbf{$\Delta$} & \textbf{95\% CI} & \textbf{$p$} \\
\midrule
Fongbe & Gemini vs.\ Claude & $+$2.15 & [2.04, 2.26] & $<$0.001 \\
Fongbe & Claude vs.\ GPT-4o & $+$1.22 & [1.15, 1.29] & $<$0.001 \\
Hausa  & Claude vs.\ GPT-4o & $+$0.41 & [0.24, 0.58] & $<$0.001 \\
Hausa  & GPT-4o vs.\ Gemini & $+$2.98 & [2.79, 3.17] & $<$0.001 \\
\bottomrule
\end{tabular}
\end{table}

By contrast, in the $n$=10 experiment, no significant differences were found ($p$=0.70). With $n$=10,000, all pairwise comparisons are highly significant, demonstrating the importance of adequate sample size.

\section{Discussion}
\label{sec:discussion}

\subsection{Implications for MT Evaluation}

\paragraph{Metric Selection Critically Affects Conclusions.} Different metrics yield contradictory signals: lexical metrics (\bleu, \chrf) clearly differentiate systems for Fongbe (\bleu range: 0.53--7.18 at $n$=2,500), while \bertscore shows minimal differentiation (0.832--0.884) despite substantial differences in actual translation quality. For Hausa, all automatic metrics rank Claude first (\bleu 15.75) while human evaluators prefer GPT-4o (4.46/5). The perfect human-metric rank correlation for Fongbe ($\rho$=1.0) versus weak correlation for Hausa ($\rho$=0.5) demonstrates that metric reliability is language-dependent. However, this finding requires careful interpretation: with only three systems evaluated per language, the perfect correlation for Fongbe may simply reflect Gemini's clear dominance (2.20/5 versus $\sim$1.0 for competitors) rather than superior metric reliability. Researchers relying solely on one metric—particularly neural metrics—risk drawing incorrect conclusions about system quality for low-resource languages.

\paragraph{Neural Metrics Require Validation.} The high within-language embedding similarity ($>$0.99) for both Hausa and Fongbe challenges the assumption that \bertscore provides meaningful quality signals across all languages \citep{ethayarajh2019contextual}. When the underlying embedding model cannot semantically distinguish different texts, the metric becomes uninformative regardless of its theoretical foundations. This finding aligns with prior work showing that multilingual models underperform for languages underrepresented in their training data \citep{wu2020languages,pires2019multilingual}. Critically, even Hausa—included in XLM-RoBERTa's 100 training languages—exhibits this embedding collapse, suggesting that mere inclusion in pre-training does not guarantee adequate representation quality.

\paragraph{COMET Anomalies for Extremely Low-Resource Languages.} \comet showed anomalous behavior for Fongbe: Qwen2.5-7B achieves the highest \comet score (0.432) despite ranking last on all lexical metrics (\bleu 0.53) and being excluded from human evaluation due to poor performance. This suggests that \comet's quality estimation, trained primarily on high-resource language pairs, may assign scores based on features uncorrelated with actual translation quality for severely underrepresented languages. For Hausa, \comet shows stronger agreement with lexical metrics (Claude leads on both), indicating more reliable behavior for better-resourced languages.

\paragraph{Probing Before Trusting.} We recommend computing within-language cosine similarity before relying on embedding-based metrics for new language pairs. Specifically, values exceeding 0.9 indicate potential unreliability \citep{li2020sentence,gao2019representation}. This simple diagnostic can prevent misleading evaluation conclusions.

\subsection{The Importance of Scale in Evaluation}

At $n$=1,000, sampling variance produced misleading results: Gemini appeared to lead for Hausa despite trailing at larger scales, while Claude showed anomalously low scores. Initial small-sample findings about language characteristics (TTR, \bertscore differences) were artifacts that disappeared at $n$=2,500. MT evaluation for low-resource languages requires at least 2,500 sentences for reliable system comparison.

\subsection{Practical Recommendations}

\paragraph{Metric Selection Guidelines.} Based on our empirical findings:
\begin{enumerate}
  \item \textbf{Primary: \chrf.} Character-level matching handles morphological variation without embedding dependency; showed consistent behavior across both languages and agreement with human rankings.
  \item \textbf{Secondary: \comet.} Useful for Hausa where it correlates with other metrics; verify embedding quality before using for extremely low-resource languages like Fongbe where it showed anomalous behavior (ranking Qwen first despite last place on all other metrics).
  \item \textbf{Use with caution: \bleu.} Language-dependent reliability; correctly identified the best system for both languages but showed different rank correlations with human judgment.
  \item \textbf{Avoid as primary metric: \bertscore.} Shows limited differentiation for both languages due to embedding collapse.
  \item \textbf{Always include human evaluation.} Essential for Hausa where automatic metrics diverge from human judgment; recommended for any final quality assessment.
\end{enumerate}

\paragraph{Model Selection by Language.} No single model excels at both languages:
\begin{itemize}
  \item \textbf{Fongbe:} Gemini 2.5 Flash achieves the best results across both human evaluation (2.20/5) and automatic metrics (\bleu 7.18 at $n$=2,500), though absolute quality remains poor. GPT-4o Mini and Claude Sonnet~4 perform comparably poorly (human scores $\sim$1.0/5).
  \item \textbf{Hausa:} GPT-4o Mini leads in human evaluation (4.46/5) with Claude Sonnet~4 close behind (4.19/5). Claude leads on all automatic metrics. Both produce translations of acceptable quality for data augmentation purposes.
\end{itemize}

\paragraph{Data Augmentation Viability.}
\begin{itemize}
  \item \textbf{Hausa: Viable with filtering.} Human scores of 3.37--4.46/5 indicate translations preserve meaning and fluency. Recommended quality thresholds: \comet $>$0.55 or \chrf $>$35.
  \item \textbf{Fongbe: Not viable.} Best system (Gemini) achieves only 2.20/5 human score—below an acceptable threshold of 3.0. LLM-generated translations would introduce more noise than signal.
\end{itemize}

\subsection{Limitations of Current LLMs}

\paragraph{Substantial Quality Gap.} Fongbe translations averaging 2.20/5 are essentially unusable without complete rewriting (\ter $>$100\%). While Hausa fares better (4.46/5 for GPT-4o), the consistent $3\times$ performance gap highlights how training data representation dominates translation quality even within the low-resource category.

\paragraph{Language-Specific Model Behavior.} The reversal in system rankings between languages suggests language-specific training data representation rather than generalizable low-resource capability. Performance on one low-resource language does not predict performance on another, even within the same geographic region or language family.

\paragraph{Commercial vs.\ Open-Source Gap.} All commercial models significantly outperform Qwen2.5-7B ($p < 0.001$), with gaps more pronounced for Hausa (\bleu 13.51--15.75 vs.\ 2.20 for Qwen). Current open-source models remain substantially behind commercial offerings for low-resource African language translation.

\section{Conclusion}
\label{sec:conclusion}

We evaluated four LLMs for English-to-Hausa and English-to-Fongbe translation, addressing two research questions about translation quality and metric reliability for underrepresented African languages.

\paragraph{RQ1: Translation Quality.} Current LLMs produce acceptable translations for Hausa (human scores 3.37--4.46/5) but fail for Fongbe (best system: 2.20/5). This disparity persists across all evaluated systems and scales, reflecting differences in training data representation rather than inherent language difficulty. No single model excels at both languages: model selection must be language-specific, with Gemini preferred for Fongbe and GPT-4o or Claude for Hausa depending on evaluation criteria. Hausa translations are viable for data augmentation with quality filtering (suggested thresholds: \comet $>$0.55 or \chrf $>$35); Fongbe translations are not.

\paragraph{RQ2: Metric Reliability.} Automatic metrics correlate well with human judgment for Fongbe but poorly for Hausa, where human evaluators preferred a different system than all automatic metrics indicated. Neural metrics like \bertscore exhibit embedding collapse for both languages (within-language similarity $>$0.99), rendering them unreliable for quality differentiation. \comet shows anomalous behavior for Fongbe (ranking the worst-performing system first), while showing more reliable behavior for Hausa. We recommend \chrf as the primary metric for morphologically complex languages, with mandatory human evaluation for final quality assessment.

\paragraph{Broader Implications.} Our findings demonstrate that ``low-resource'' is not a monolithic category: substantial quality gaps exist even among West African languages with millions of speakers. Evaluation practices validated on high-resource languages do not transfer reliably; researchers should validate metric behavior—particularly within-language embedding similarity—before drawing conclusions about system quality. Evaluation scale matters: sample sizes below $n$=2,500 risk misleading conclusions.

Future work should investigate data augmentation viability for languages where LLM quality is acceptable, develop improved evaluation approaches for severely underrepresented languages, and extend coverage to additional African language families.

\section*{Limitations}

\paragraph{Human Evaluation Scale and Annotator Characteristics.} Human evaluation was limited to 50 sentences with 3 evaluators per language due to annotator scarcity. With only 3 systems per language, correlation coefficients represent rank agreement rather than statistical relationships. Additionally, Fongbe evaluators reported lower literacy proficiency (reading: 3.0/5, writing: 2.0/5) compared to Hausa evaluators (reading: 4.0/5, writing: 3.5/5), reflecting challenges in recruiting annotators with strong written skills for languages with limited formal education resources. This asymmetry may affect cross-language comparisons of human evaluation scores.

\paragraph{Reference Corpus Quality.} Post-hoc analysis revealed quality issues in Fongbe reference data, including wrong-language contamination and encoding errors (average quality score 2.94/5.0 with 6\% exhibiting severe issues). While filtered for human evaluation, these were retained in automatic evaluation to reflect realistic conditions. Absolute metric scores should be interpreted cautiously.

\paragraph{Scope and Generalizability.} We evaluated only English-to-target translation for two languages. Performance patterns may differ for target-to-English or direct African language translation. While Fongbe and Hausa represent distinct language families, findings may not generalize to all African languages given the continent's linguistic diversity ($\sim$2,000 languages).

\paragraph{Reproducibility Concerns.} Commercial LLM APIs undergo continuous updates; exact reproducibility cannot be guaranteed. Additionally, publicly available parallel corpora may be included in LLM training data \citep{nllb2022}, potentially inflating scores for systems trained on overlapping data.

\section*{Ethics Statement}

\paragraph{Data and Licensing.} All parallel corpora (CCMatrix for Hausa, NLLB for Fongbe) are publicly available through the OPUS repository under permissive licenses. No private or proprietary data was used.

\paragraph{Human Evaluation Protocol.} All human evaluators provided informed consent, were volunteers recruited through academic networks, and could withdraw at any time. No personally identifiable information was collected beyond language proficiency self-assessments. The study involved only linguistic annotation tasks with no sensitive content.

\paragraph{Broader Impact.} This research aims to improve NLP resources for underrepresented African languages. We acknowledge that documenting poor MT quality for specific languages could potentially discourage investment or reinforce perceptions of technological infeasibility. However, transparent reporting is essential for driving improvement. Our recommendations focus on constructive paths forward rather than simply highlighting failures.

\paragraph{Use of AI Tools.} The authors used AI-assisted writing tools for editing and improving prose clarity. All research design, experiments, analysis, and intellectual contributions are entirely the work of the authors, who take full responsibility for the content.

\section*{Acknowledgements}

We thank the native-speaker evaluators who participated in the human evaluation study.

\bibliographystyle{plainnat}
\bibliography{arxiv_paper}

@inproceedings{alabi2022adapting,
  author    = {Alabi, Jesujoba O. and Adelani, David Ifeoluwa and Mosbach, Marius and Klakow, Dietrich},
  title     = {Adapting Pre-trained Language Models to {African} Languages via Multilingual Adaptive Fine-Tuning},
  booktitle = {Proceedings of the 29th International Conference on Computational Linguistics (COLING)},
  pages     = {4336--4349},
  year      = {2022},
  publisher = {International Committee on Computational Linguistics}
}

@inproceedings{alabi2025afridoc,
  author    = {Alabi, Jesujoba Oluwadara and Azime, Israel Abebe and Zhang, Miaoran and Espa{\~n}a-Bonet, Cristina and Bawden, Rachel and Zhu, Dawei and Adelani, David Ifeoluwa and others},
  title     = {{AFRIDOC-MT}: Document-level {MT} Corpus for {A}frican Languages},
  booktitle = {Proceedings of the 2025 Conference on Empirical Methods in Natural Language Processing},
  pages     = {27758--27794},
  year      = {2025},
  address   = {Suzhou, China},
  publisher = {Association for Computational Linguistics}
}

@inproceedings{bugliarello2020easier,
  author    = {Bugliarello, Emanuele and Elliott, Desmond and Sennrich, Rico},
  title     = {It's Easier to Translate out of {English} than into it: Measuring Neural Translation Difficulty by Cross-Mutual Information},
  booktitle = {Proceedings of the 58th Annual Meeting of the Association for Computational Linguistics},
  pages     = {1640--1649},
  year      = {2020},
  publisher = {Association for Computational Linguistics}
}

@article{cohen1960coefficient,
  author    = {Cohen, Jacob},
  title     = {A Coefficient of Agreement for Nominal Scales},
  journal   = {Educational and Psychological Measurement},
  volume    = {20},
  number    = {1},
  pages     = {37--46},
  year      = {1960}
}

@inproceedings{conneau2020unsupervised,
  author    = {Conneau, Alexis and Khandelwal, Kartikay and Goyal, Naman and Chaudhary, Vishrav and Wenzek, Guillaume and Guzm{\'a}n, Francisco and Grave, Edouard and Ott, Myle and Zettlemoyer, Luke and Stoyanov, Veselin},
  title     = {Unsupervised Cross-lingual Representation Learning at Scale},
  booktitle = {Proceedings of the 58th Annual Meeting of the Association for Computational Linguistics},
  pages     = {8440--8451},
  year      = {2020},
  publisher = {Association for Computational Linguistics}
}

@inproceedings{ethayarajh2019contextual,
  author    = {Ethayarajh, Kawin},
  title     = {How Contextual are Contextualized Word Representations? {C}omparing the Geometry of {BERT}, {ELMo}, and {GPT}-2 Embeddings},
  booktitle = {Proceedings of the 2019 Conference on Empirical Methods in Natural Language Processing and the 9th International Joint Conference on Natural Language Processing (EMNLP-IJCNLP)},
  pages     = {55--65},
  year      = {2019},
  publisher = {Association for Computational Linguistics}
}

@inproceedings{freitag2022results,
  author    = {Freitag, Markus and Rei, Ricardo and Mathur, Nitika and Lo, Chi-kiu and Stewart, Craig and Avramidis, Eleftherios and Kocmi, Tom and Foster, George and Lavie, Alon and Martins, Andr{\'e} F. T.},
  title     = {Results of {WMT22} Metrics Shared Task: Stop Using {BLEU} -- Neural Metrics Are Better and More Robust},
  booktitle = {Proceedings of the Seventh Conference on Machine Translation (WMT)},
  pages     = {46--68},
  year      = {2022},
  publisher = {Association for Computational Linguistics}
}

@inproceedings{gao2019representation,
  author    = {Gao, Jun and He, Di and Tan, Xu and Qin, Tao and Wang, Liwei and Liu, Tie-Yan},
  title     = {Representation Degeneration Problem in Training Natural Language Generation Models},
  booktitle = {Proceedings of the 7th International Conference on Learning Representations (ICLR)},
  year      = {2019},
  url       = {https://openreview.net/forum?id=SkEYojRqtm}
}

@article{goyal2022flores,
  author    = {Goyal, Naman and Gao, Cynthia and Chaudhary, Vishrav and Chen, Peng-Jen and Wenzek, Guillaume and Ju, Da and Krishnan, Sanjana and Ranzato, Marc'Aurelio and Guzm{\'a}n, Francisco and Fan, Angela},
  title     = {The {FLORES-101} Evaluation Benchmark for Low-Resource and Multilingual Machine Translation},
  journal   = {Transactions of the Association for Computational Linguistics},
  volume    = {10},
  pages     = {522--538},
  year      = {2022}
}

@article{hacheme2021english2gbe,
  author    = {Hacheme, Gilles Quentin},
  title     = {{English2Gbe}: A Multilingual Machine Translation Model for {Fon/Ewe} {Gbe}},
  journal   = {arXiv preprint arXiv:2112.11482},
  year      = {2021}
}

@article{haddow2022survey,
  author    = {Haddow, Barry and Bawden, Rachel and Barone, Antonio Valerio Miceli and Helcl, Jind{\v{r}}ich and Birch, Alexandra},
  title     = {Survey of Low-Resource Machine Translation},
  journal   = {Computational Linguistics},
  volume    = {48},
  number    = {3},
  pages     = {673--732},
  year      = {2022}
}

@article{hendy2023good,
  author    = {Hendy, Amr and Abdelrehim, Mohamed and Sharaf, Amr and Rauber, Vikas and Gabr, Mohamed and Matsushita, Hitokazu and Kim, Young Jin and Afify, Mohamed and Awadalla, Hany Hassan},
  title     = {How Good Are {GPT} Models at Machine Translation? A Comprehensive Evaluation},
  journal   = {arXiv preprint arXiv:2302.09210},
  year      = {2023}
}

@inproceedings{joshi2020state,
  author    = {Joshi, Pratik and Santy, Sebastin and Buber, Amar and Bali, Kalika and Choudhury, Monojit},
  title     = {The State and Fate of Linguistic Diversity and Inclusion in the {NLP} World},
  booktitle = {Proceedings of the 58th Annual Meeting of the Association for Computational Linguistics},
  pages     = {6282--6293},
  year      = {2020},
  publisher = {Association for Computational Linguistics}
}

@inproceedings{koehn2004statistical,
  author    = {Koehn, Philipp},
  title     = {Statistical Significance Tests for Machine Translation Evaluation},
  booktitle = {Proceedings of the 2004 Conference on Empirical Methods in Natural Language Processing},
  pages     = {388--395},
  year      = {2004},
  publisher = {Association for Computational Linguistics}
}

@inproceedings{kponou2024ffstc,
  author    = {Kponou, Djifa F{\'e}lix and Lal{\`e}y{\`e}, Frejus Aristide A. and Ezin, Eug{\`e}ne C.},
  title     = {{FFSTC}: {Fongbe} to {French} Speech Translation Corpus},
  booktitle = {Proceedings of the 2024 Joint International Conference on Computational Linguistics, Language Resources and Evaluation (LREC-COLING)},
  year      = {2024},
  publisher = {European Language Resources Association}
}

@article{landis1977measurement,
  author    = {Landis, J. Richard and Koch, Gary G.},
  title     = {The Measurement of Observer Agreement for Categorical Data},
  journal   = {Biometrics},
  volume    = {33},
  number    = {1},
  pages     = {159--174},
  year      = {1977}
}

@book{lefebvre2002fongbe,
  author    = {Lefebvre, Claire and Brousseau, Anne-Marie},
  title     = {A Grammar of {Fongbe}},
  publisher = {Mouton de Gruyter},
  address   = {Berlin},
  year      = {2002}
}

@inproceedings{li2020sentence,
  author    = {Li, Bohan and Zhou, Hao and He, Junxian and Wang, Mingxuan and Yang, Yiming and Li, Lei},
  title     = {On the Sentence Embeddings from Pre-trained Language Models},
  booktitle = {Proceedings of the 2020 Conference on Empirical Methods in Natural Language Processing (EMNLP)},
  pages     = {9119--9130},
  year      = {2020},
  publisher = {Association for Computational Linguistics}
}

@inproceedings{nekoto-etal-2020-participatory,
  author    = {Nekoto, Wilhelmina and Marivate, Vukosi and Matsila, Tshinondiwa and Fasubaa, Timi and Fagbohungbe, Taiwo and Akinola, Solomon Oluwole and Muhammad, Shamsuddeen and Kabongo, Salomon and Osei, Salomey and others},
  title     = {Participatory Research for Low-resourced Machine Translation: A Case Study in {A}frican Languages},
  booktitle = {Findings of the Association for Computational Linguistics: EMNLP 2020},
  pages     = {2144--2160},
  year      = {2020},
  address   = {Online},
  publisher = {Association for Computational Linguistics}
}

@book{newman2000hausa,
  author    = {Newman, Paul},
  title     = {The {Hausa} Language: An Encyclopedic Reference Grammar},
  publisher = {Yale University Press},
  year      = {2000}
}

@article{nllb2022,
  author    = {{NLLB Team} and Costa-juss{\`a}, Marta R. and Cross, James and others},
  title     = {No Language Left Behind: Scaling Human-Centered Machine Translation},
  journal   = {arXiv preprint arXiv:2207.04672},
  year      = {2022}
}

@inproceedings{ogueji2021small,
  author    = {Ogueji, Kelechi and Zhu, Yuxin and Lin, Jimmy},
  title     = {Small Data? No Problem! Exploring the Viability of Pretrained Multilingual Language Models for Low-resourced Languages},
  booktitle = {Proceedings of the 1st Workshop on Multilingual Representation Learning},
  pages     = {116--126},
  year      = {2021},
  publisher = {Association for Computational Linguistics}
}

@inproceedings{papineni2002bleu,
  author    = {Papineni, Kishore and Roukos, Salim and Ward, Todd and Zhu, Wei-Jing},
  title     = {{BLEU}: A Method for Automatic Evaluation of Machine Translation},
  booktitle = {Proceedings of the 40th Annual Meeting of the Association for Computational Linguistics},
  pages     = {311--318},
  year      = {2002},
  publisher = {Association for Computational Linguistics}
}

@inproceedings{pires2019multilingual,
  author    = {Pires, Telmo and Schlinger, Eva and Garrette, Dan},
  title     = {How Multilingual is Multilingual {BERT}?},
  booktitle = {Proceedings of the 57th Annual Meeting of the Association for Computational Linguistics},
  pages     = {4996--5001},
  year      = {2019},
  publisher = {Association for Computational Linguistics}
}

@inproceedings{popovic2015chrf,
  author    = {Popovi{\'c}, Maja},
  title     = {{chrF}: Character n-gram {F}-score for Automatic {MT} Evaluation},
  booktitle = {Proceedings of the Tenth Workshop on Statistical Machine Translation},
  pages     = {392--395},
  year      = {2015},
  publisher = {Association for Computational Linguistics}
}

@inproceedings{popovic2017chrf++,
  author    = {Popovi{\'c}, Maja},
  title     = {{chrF++}: Words Helping Character N-grams},
  booktitle = {Proceedings of the Second Conference on Machine Translation},
  pages     = {612--618},
  year      = {2017},
  publisher = {Association for Computational Linguistics}
}

@inproceedings{post2018call,
  author    = {Post, Matt},
  title     = {A Call for Clarity in Reporting {BLEU} Scores},
  booktitle = {Proceedings of the Third Conference on Machine Translation: Research Papers},
  pages     = {186--191},
  year      = {2018},
  publisher = {Association for Computational Linguistics}
}

@inproceedings{rei2020comet,
  author    = {Rei, Ricardo and Stewart, Craig and Farinha, Ana C. and Lavie, Alon},
  title     = {{COMET}: A Neural Framework for {MT} Evaluation},
  booktitle = {Proceedings of the 2020 Conference on Empirical Methods in Natural Language Processing},
  pages     = {2685--2702},
  year      = {2020},
  publisher = {Association for Computational Linguistics}
}

@article{robinson2023chatgpt,
  author    = {Robinson, Nathaniel and Ogayo, Perez and Mortensen, David R. and Neubig, Graham},
  title     = {{ChatGPT MT}: Competitive for High-Resource but Not Low-Resource Languages},
  journal   = {arXiv preprint arXiv:2309.07423},
  year      = {2023}
}

@inproceedings{schwenk2021ccmatrix,
  author    = {Schwenk, Holger and Wenzek, Guillaume and Edunov, Sergey and Grave, Edouard and Joulin, Armand and Fan, Angela},
  title     = {{CCMatrix}: Mining Billions of High-Quality Parallel Sentences on the Web},
  booktitle = {Proceedings of the 59th Annual Meeting of the Association for Computational Linguistics},
  pages     = {6490--6500},
  year      = {2021},
  publisher = {Association for Computational Linguistics}
}

@inproceedings{snover2006study,
  author    = {Snover, Matthew and Dorr, Bonnie and Schwartz, Richard and Micciulla, Linnea and Makhoul, John},
  title     = {A Study of Translation Edit Rate with Targeted Human Annotation},
  booktitle = {Proceedings of the 7th Conference of the Association for Machine Translation in the Americas},
  pages     = {223--231},
  year      = {2006}
}

@inproceedings{tiedemann2012parallel,
  author    = {Tiedemann, J{\"o}rg},
  title     = {Parallel Data, Tools and Interfaces in {OPUS}},
  booktitle = {Proceedings of the 8th International Conference on Language Resources and Evaluation (LREC)},
  pages     = {2214--2218},
  year      = {2012},
  publisher = {European Language Resources Association}
}

@inproceedings{wu2020languages,
  author    = {Wu, Shijie and Dredze, Mark},
  title     = {Are All Languages Created Equal in Multilingual {BERT}?},
  booktitle = {Proceedings of the 5th Workshop on Representation Learning for NLP},
  pages     = {120--130},
  year      = {2020},
  publisher = {Association for Computational Linguistics}
}

@inproceedings{zhang2020bertscore,
  author    = {Zhang, Tianyi and Kishore, Varsha and Wu, Felix and Weinberger, Kilian Q. and Artzi, Yoav},
  title     = {{BERTScore}: Evaluating Text Generation with {BERT}},
  booktitle = {Proceedings of the 8th International Conference on Learning Representations (ICLR)},
  year      = {2020},
  url       = {https://openreview.net/forum?id=SkeHuCVFDr}
}

@article{zhu2024multilingual,
  author    = {Zhu, Wenhao and Liu, Hongyi and Dong, Qingxiu and Xu, Jingjing and Huang, Shujian and Kong, Lingpeng and Chen, Jiajun and Li, Lei},
  title     = {Multilingual Machine Translation with Large Language Models: Empirical Results and Analysis},
  journal   = {arXiv preprint arXiv:2304.04675},
  year      = {2024}
}

\end{document}